\documentclass[9pt,conference,compsocconf]{IEEEtran}
\usepackage{float}
\usepackage{hyperref}
\usepackage{graphicx}	%
\usepackage{amsmath}
\usepackage[T1]{fontenc}

\begin{document}
	\title{Model Generalization: A Sharpness Aware Optimization Perspective}

	\author{\IEEEauthorblockN{Jozef Marus Coldenhoff$^*$}
		\textit{School of Computer and Communication Sciences}\\
		EPFL \\
		\and
		\IEEEauthorblockN{Chengkun Li$^*$}
		\textit{School of Engineering}\\
		EPFL \\
		\and
		\IEEEauthorblockN{Yurui Zhu$^*$} 
		\textit{College of Humanities}\\
		EPFL \\
	}
	
	\maketitle
	\def\thefootnote{*}\footnotetext{These authors contributed equally to this work}\def\thefootnote{\arabic{footnote}}

	\begin{abstract}
		Sharpness-Aware Minimization (SAM) and adaptive sharpness-aware minimization (ASAM) aim to improve the model generalization. And in this project, we proposed three experiments to valid their generalization from the sharpness aware perspective. And our experiments show that sharpness aware-based optimization techniques could help to provide models with strong generalization ability. Our experiments also show that ASAM could improve the generalization performance on un-normalized data, but further research is needed to confirm this.
	\end{abstract}

	\begin{IEEEkeywords}
		Machine Learning, Optimization, Loss sharpness, SAM, ASAM.
	\end{IEEEkeywords}
	
	\section{Introduction}
	
	In recent years, the generalization of deep neural networks has steadily risen to prominence as an important topic in modern machine learning, plenty of works have been done to address the limitations of pure optimization. Better model generalization means more stable results when dealing with unobserved data in real-world applications. Empirical evidence has shown that the shape of the loss function affects how well the models generalize; convergence at a flatter minima is more likely to lead to better generalization. 
	
	Stochastic gradient descent (SGD) is known for finding a flat minima that tends to generalize well. Xie and Sato \textit{et al.} \cite{xie_diffusion_2021} theoretically proved that SGD favors flat minima exponentially due to the Hessian-dependent covariance of stochastic gradient noise. Furthermore, Jastrzebski \textit{et al.} \cite{jastrzebski_finding_2018} theoretically and empirically showed that using a large learning rate and/or small batch size steers SGD towards flatter minima. 
	
	Foret and Kleiner \textit{et al.} \cite{foret_sharpness-aware_2021} introduced Sharpness-Aware Minimization (SAM) to improve generalization by simultaneously minimizing both loss value and loss sharpness. Subsequently, Kwon and Kim \textit{et al.} \cite{kwon_asam_2021} proposed adaptive sharpness-aware minimization (ASAM), which adaptively adjusts maximization regions thus acting uniformly under parameter re-scaling. Since the advent of SAM, many studies have demonstrated that SAM combined with different optimization methods can all improve the generalization. \cite{andriushchenko_understanding_2021,anand_towards_2022,na_train_2022,bahri2021sharpness}. However, there are not many experiments to validate the performance of ASAM.  
	
	In this project, we aim to evaluate how the SAM and ASAM optimizer perform on a Computer Vision image classification task, and whether SGD using SAM and ASAM can truly beat vanilla SGD in terms of generalization performance.
	
	\section{Methods}
	\label{sec:method}
	
	The Sharpness-Aware Minimization (SAM) performs one more step of gradient ascent to approximately determine the worst-case weight perturbation before updating the weights \cite{foret_sharpness-aware_2021}. This is done, because in the paper they derive that adding a sharpness minimization objective to the loss function can be optimized by evaluating the gradient or the loss at a point in parameter space within an $\epsilon$-ball around the current parameters, and adding that evaluated gradient to the current parameters. This has as an intended effect that the network balances lowering the training loss, and its sharpness.
	More specifically, \cite{foret_sharpness-aware_2021} defines the sharpness of loss function as Equation \ref{eq:1}.
	\begin{equation}
		\max_{\|\boldsymbol{\epsilon}\|{2} \leq \rho} L_{S}(\mathbf{w}+\boldsymbol{\epsilon})-L_{S}(\mathbf{w})
		\label{eq:1}
	\end{equation}
	
	And the sharpness-aware minimization can be defined as
	the following min-max optimization:
	
	\begin{equation}
		\min {\mathbf{w}} \max _{\|\boldsymbol{\epsilon}\|{p} \leq \rho} L_{S}(\mathbf{w}+\boldsymbol{\epsilon})+\frac{\lambda}{2}\|\mathbf{w}\|_{2}^{2}
	\end{equation}
	
	And the weights can be updated after the \(\epsilon\) is calculated, using Equation \ref{eq:3}:
	\begin{equation}
		\left\{\begin{array}{l}
			\boldsymbol{\epsilon}_{t}=\rho \frac{\nabla L{S}\left(\mathbf{w}_{t}\right)}{\left\|\nabla L{S}\left(\mathbf{w}_{t}\right)\right\|{2}} \\
			\mathbf{w}_{t+1}=\mathbf{w}_{t}-\alpha_{t}\left(\nabla L_{S}\left(\mathbf{w}_{t}+\boldsymbol{\epsilon}_{t}\right)+\lambda \mathbf{w}_{t}\right)
		\end{array}\right.
		\label{eq:3}
	\end{equation}

	However, \cite{kwon_asam_2021} note that as was shown by \cite{https://doi.org/10.48550/arxiv.1703.04933} a rectifier neural network can be arbitrarily reparameterized in such a way that the function of the network stays the same, while changing the sharpness of the loss landscape. \cite{kwon_asam_2021} then suggest that rescaling the $\epsilon$-ball to match the scale of the parameters could improve the performance of the SAM algorithm by removing the scale-dependence from the loss maximization step. This resulted in the ASAM optimizer, where the maximization problem is changed to:
	
	\begin{equation}
		\max_{\|\boldsymbol{T_w^{-1}\epsilon}\|{2} \leq \rho} L_{S}(\mathbf{w}+\boldsymbol{\epsilon})-L_{S}(\mathbf{w})
		\label{eq:4}
	\end{equation}
	Where $T_w^{-1}$ is a rescaling operator defined by the magnitudes of the parameters in fully connected and convolution layers.
	
	The final weight update is the calculated as: 
	
	\begin{equation}
		\left\{\begin{array}{l}
			\boldsymbol{\epsilon}_{t}=\rho \frac{T^2_w \nabla L{B}\left(\mathbf{w}_{t}\right)}{\left\|T_w\nabla L{S}\left(\mathbf{w}_{t}\right)\right\|{2}} \\
			\mathbf{w}_{t+1}=\mathbf{w}_{t}-\alpha\left(\nabla L_{B}\left(\mathbf{w}_{t}+\boldsymbol{\epsilon}\right)+\lambda \mathbf{w}_{t}\right)
		\end{array}\right.
	\end{equation}
	
	We carried out the following experiments to discover model's generalization from the sharpness aware perspective.
	
	\section{Experiments}
	\subsection{Experiment A}\label{sec:exp1}
	
	The aim of our first experiment is to test the generalization ability of model trained with \textit{Sharpness Aware Minimization }(SAM) and vanilla \textit{Stochastic Gradient Descent} (SGD) on image classification task at testing time. To this end, we trained ResNet-18 on CIFAR-10 for 100 epochs using SAM and SGD. For SAM, it needs a base optimizer, so we use a SGD with the same configuration with the SGD we compared to in order to ensure fairness. The SGD is trained with 0 momentum and constant learning rates of 0.01 (the learning rate is selected by grid search with the goal of balancing training time and training stability) with batch size of 128, for image classification task we use Cross Entropy Loss. We measured the classification accuracy of models trained with these two optimizers on test data-set to determine their generalization ability.
	
	Additionally, since SAM updates twice per training step, so we compare the testing accuracy of SAM at first 50 epochs with the testing accuracy of SGD at first 100 epochs. We report the results in Table~\ref{tab:exp1} which are measured as average across 3 seeds. 
	\begin{table}[h] 
		\scriptsize
		\centering
		\begin{tabular}{|l|l|l|}
			\hline
			Seed & Method & Testing accuracy \\ \hline
			20   & SAM   & \textbf{71.22}    \\ \cline{2-3} 
			& SGD   & 65.15             \\ \hline
			30   & SAM   & \textbf{70.77}    \\ \cline{2-3} 
			& SGD   & 64.03             \\ \hline
			40   & SAM   & \textbf{70.98}    \\ \cline{2-3} 
			& SGD   & 64.17             \\ \hline
		\end{tabular}
		\caption{Result of experiment 1: comparing the testing accuracy of models trained with SGD/SAM.}
		\label{tab:exp1} 
	\end{table}
	
	From the result we can see that with every random initialization, SAM has a significantly higher testing accuracy than vanilla SGD.

	\subsection{Experiment B}
	According to Andriushchenko \textit{et al.} \cite{andriushchenko_understanding_2021}, using smaller batch sizes steers SGD toward flatter minima, and in contrast, a large batch size theoretically makes it harder for SGD harder to escape from sharper minima due to less noise in the gradient. In the second experiment, we will explore whether SAM helps to mitigate the poor generalization associated with larger batch sizes of SGD and see if it can help generalize better in the test time by finding a flatter minimum. We here trained the ResNet-18 on CIFAR-10 for 100 epochs using both vanilla SGD and SAM with the same configuration in Section \ref{sec:exp1}.
	
\begin{figure}[h]
		\centering
		\includegraphics[width=0.45\textwidth]{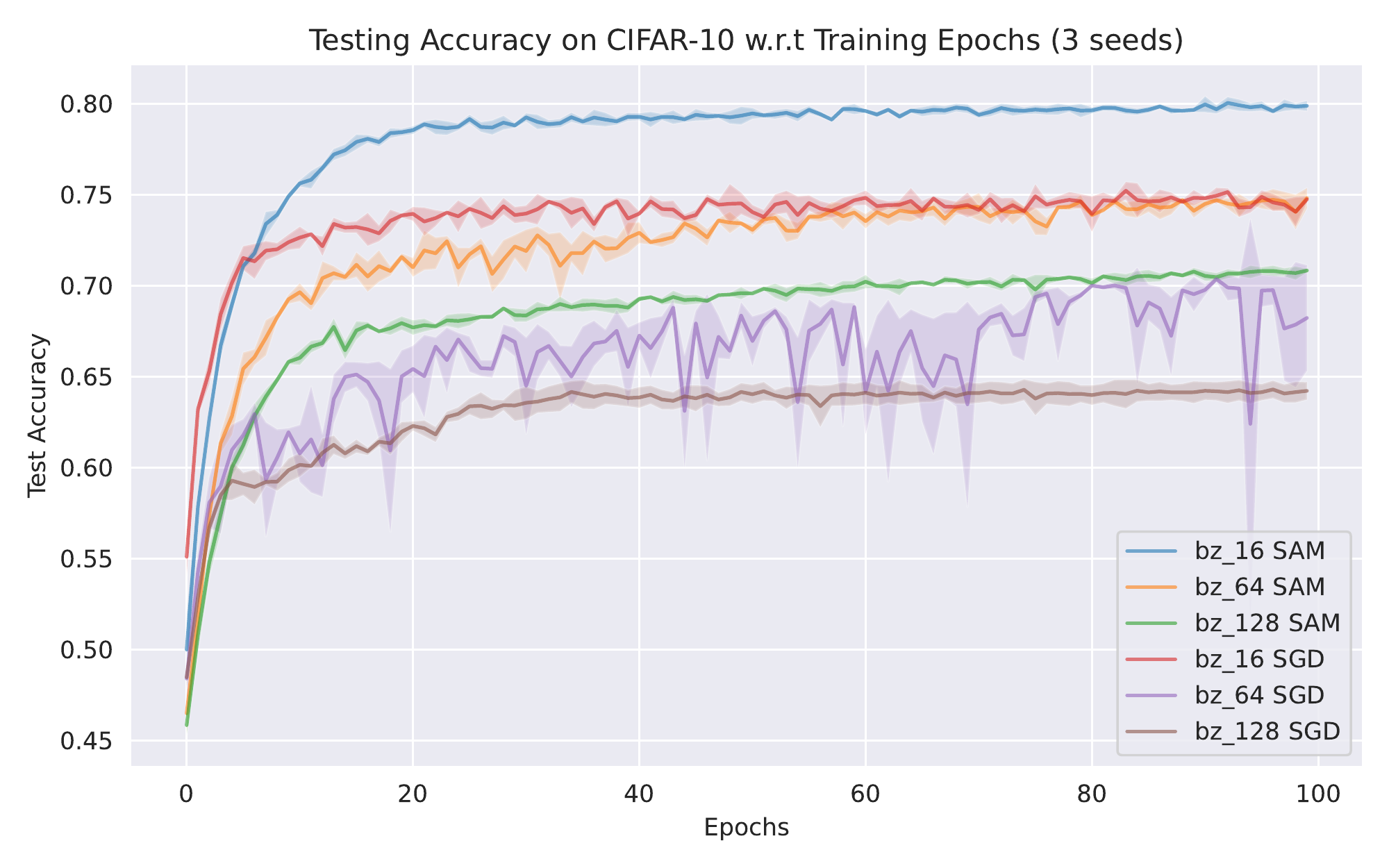}
		\caption{Testing accuracy on CIFAR-10 with SAM/SGD trained with different batch sizes and initial weights.}
		\label{fig:acc}
	\end{figure}

\begin{figure}[h]
		\centering
		\includegraphics[width=.45\textwidth]{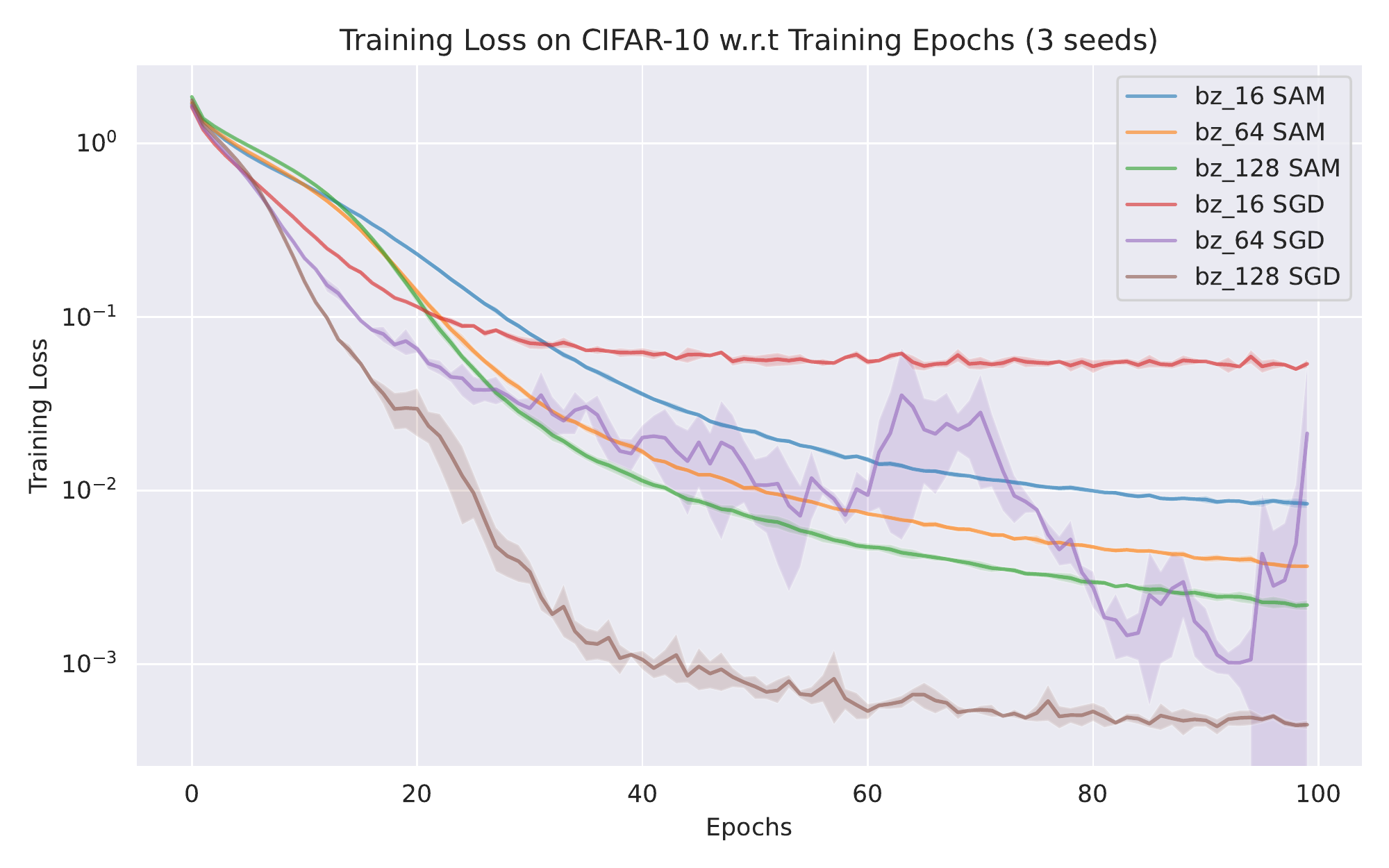}
		\caption{Training loss on CIFAR-10 with SAM/SGD trained with different batch sizes and initial weights.}
		\label{fig:loss}
	\end{figure}

	We report the testing accuracy w.r.t training epochs and training loss (in log scale) w.r.t training epochs in Figure \ref{fig:acc} and Figure \ref{fig:loss}. All curves are averaged across 3 seeds with standard deviation plotted in light colors.
	
	\textit{\textbf{Testing Accuracy Interpretation}:} From Figure \ref{fig:acc}, we can observe that models trained with smaller batch size generalize better at testing time for models trained with either SAM or SGD, which is in consistent with the observations in \cite{andriushchenko_understanding_2021}. However, from our experiments, we observe that the test accuracy of the models using SGD varies more than the test accuracy of the models using SAM across 3 seeds, this is especially evident in larger batch sizes (64 and 128), where the variance of the SAM-based model is significantly lower than that of the SGD-based model. In addition, comparing the models trained with SAM/SGD for the same batch size, the models trained with SAM achieved test accuracy at around $5\sim10\%$ higher than SGD.
	
	\textit{\textbf{Training Loss Interpretation}:} From Figure \ref{fig:loss} we can easily distinguish the loss curves of the SAM-based models from the SGD-based models by either the convergence rate or the variance of loss. Interestingly, our set of experiments shows that lower losses do not necessarily lead to higher accuracy in test time, and in addition, we can see that faster convergence in the case of SGD during training represents lower generalization performance in our experiments as well. On the other hand, we can observe a similar phenomenon for the loss curve variance, i.e., the SGD-based model has a larger variance in the loss curve compared to the SAM-based model. We interpret this result in the perspective of sharpness, this high variance may come from different weight initializations, where training with SGD may be stuck at different sharp local minima, while training with SAM leads to flatter minima, or perhaps even the same minima for different weight initializations.
	
	\subsection{Experiment C}
	
	ASAM intends to address the parameter scaling problem that SAM and other sharpness minimization-based optimizers may encounter. As it has been shown by \cite{https://doi.org/10.48550/arxiv.1703.04933} that one may arbitrarily reparameterize rectifier networks in such a way that the function of the network stays the same, while changing the flatness of the loss landscape. ASAM tackles this by rescaling the $\epsilon$-ball where the loss is maximized to reflect the different magnitudes of the parameters.
	
	We wanted to see how much this rescaling problem affects SAM in an image classification task and whether ASAM can solve the problem. Hence we conduct an experiment to test this. We design the experiment to train ResNet-18 on CIFAR-10 for 200 epochs using vanilla SGD with SAM and ASAM, in order to find whether ASAM is more robust to parameter scaling, we remove all Batch Normalization layers \cite{https://doi.org/10.48550/arxiv.1502.03167}. We conduct this experiment with the same hyperparameters as the ASAM authors use in their comparison between SAM and ASAM using CIFAR-10 and ResNet 20. The learning rate for both SAM and ASAM is set to 0.01, batch size of 128, momentum and weight decay are set to 0.9 and 0.0005, and finally, a cosine annealing learning rate scheduler \cite{https://doi.org/10.48550/arxiv.1608.03983} is used to train the networks. For this experiment, we use the standard data normalization of CIFAR-10, where the input pixels have an intensity between 0 and 1. For both SAM and ASAM, we use three seeds to initialize the network weights and average the results in the end.
	
	\begin{table}[H] 
		\centering
		\scriptsize
		\begin{tabular}{|l|l|l|l|}
			\hline
			Seed & Method & Testing accuracy & Training loss ($10^{-5}$)      \\ \hline
			1    & SAM    & \textbf{70.06}  & \textbf{8.72} \\ \cline{2-4} 
			& ASAM   & 68.65           & 27.5          \\ \hline
			2    & SAM    & 65.44           & \textbf{7.19} \\ \cline{2-4} 
			& ASAM   & \textbf{65.77}  & 25.0          \\ \hline
			3    & SAM    & 65.49           & \textbf{9.50} \\ \cline{2-4} 
			& ASAM   & \textbf{66.23}  & 24.7          \\ \hline
		\end{tabular}
		\caption{Table showing the results of SAM and ASAM trained on CIFAR-10}
		\label{Asam} 
	\end{table}
	Table \ref{Asam} shows the results of the three training runs on both SAM and ASAM. We can observe that in seed 2 and 3, the generalization performance of ASAM is slightly better than that of SAM. However, we see that the training loss of the networks trained with SAM is an order of magnitude lower than that of ASAM. It thus seems that ASAM found a local minima with a higher training loss, while being similar to SAM in terms of generalization.
	
\begin{figure}
		\centering
		\includegraphics[width = 0.45\textwidth]{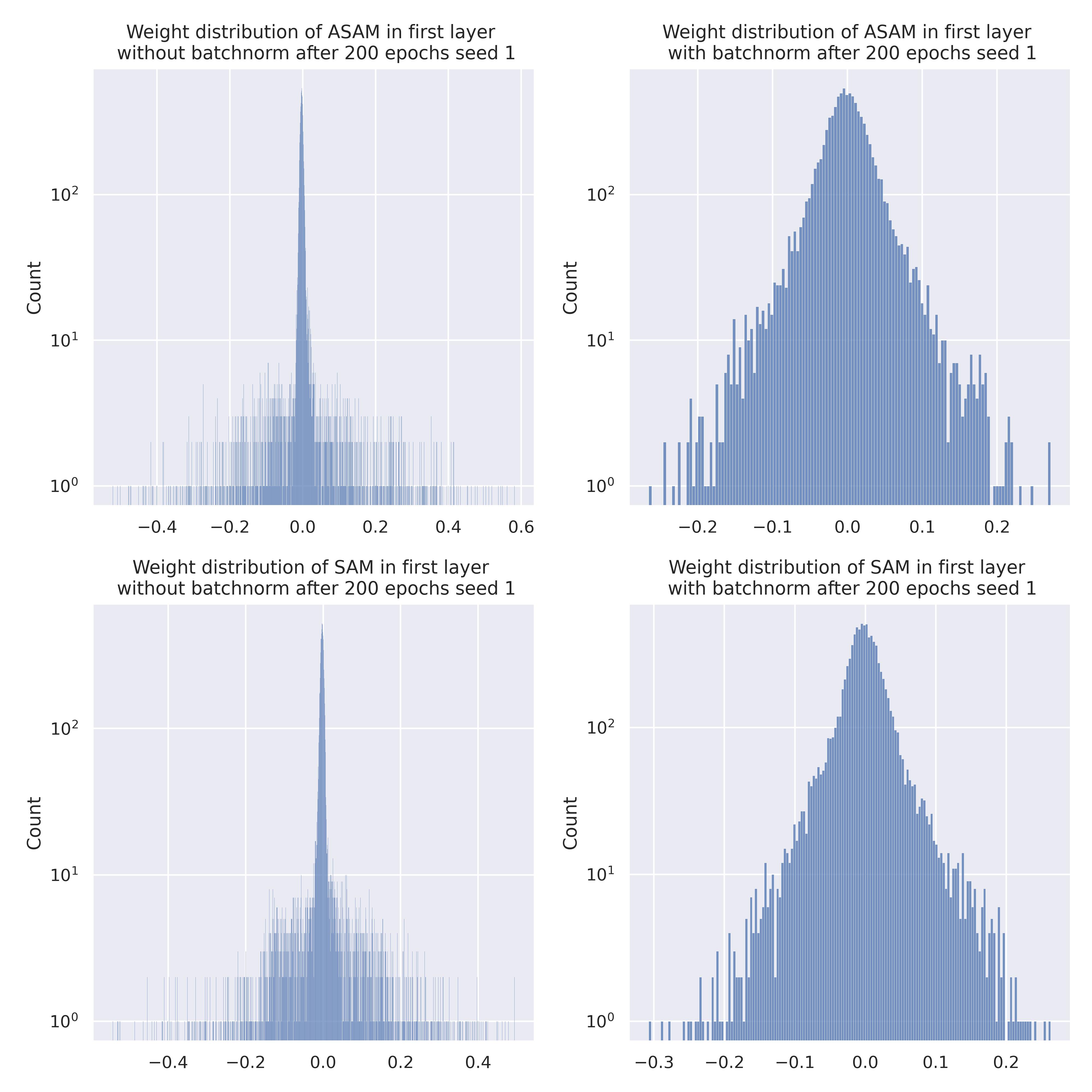}
		\caption{Weight distribution for ASAM and SAM in the first convolution layer.}
		\label{fig:weightdistr}
	\end{figure}

	In order to verify that the removal of the Batch Normalization really led to the result that weights distributing themselves over a larger range, we visualize the distribution of the weights in the first layer of the network when training with or without Batch Normalization for both SAM and ASAM. Figure \ref{fig:weightdistr} shows the distribution of the weights with the y-axis in log scale. It seems to show that the removal Batch Normalization created a more heavy tailed distribution, indicating that the weights are of different scale.
	
	Similarly to earlier experiments, we plot how the accuracy and loss evolve over epochs in Appendix \ref{appendix:asamacc} and Figure \ref{fig:asamloss} respectively. Figure \ref{fig:asamloss} shows why we observed a difference in the training loss between SAM and ASAM. We observe that in the late stages of training, the SAM optimizer seems to descend further than the ASAM, while keeping similar testing accuracy. 
	
	\begin{figure}
		\centering
		\includegraphics[width = 0.45\textwidth]{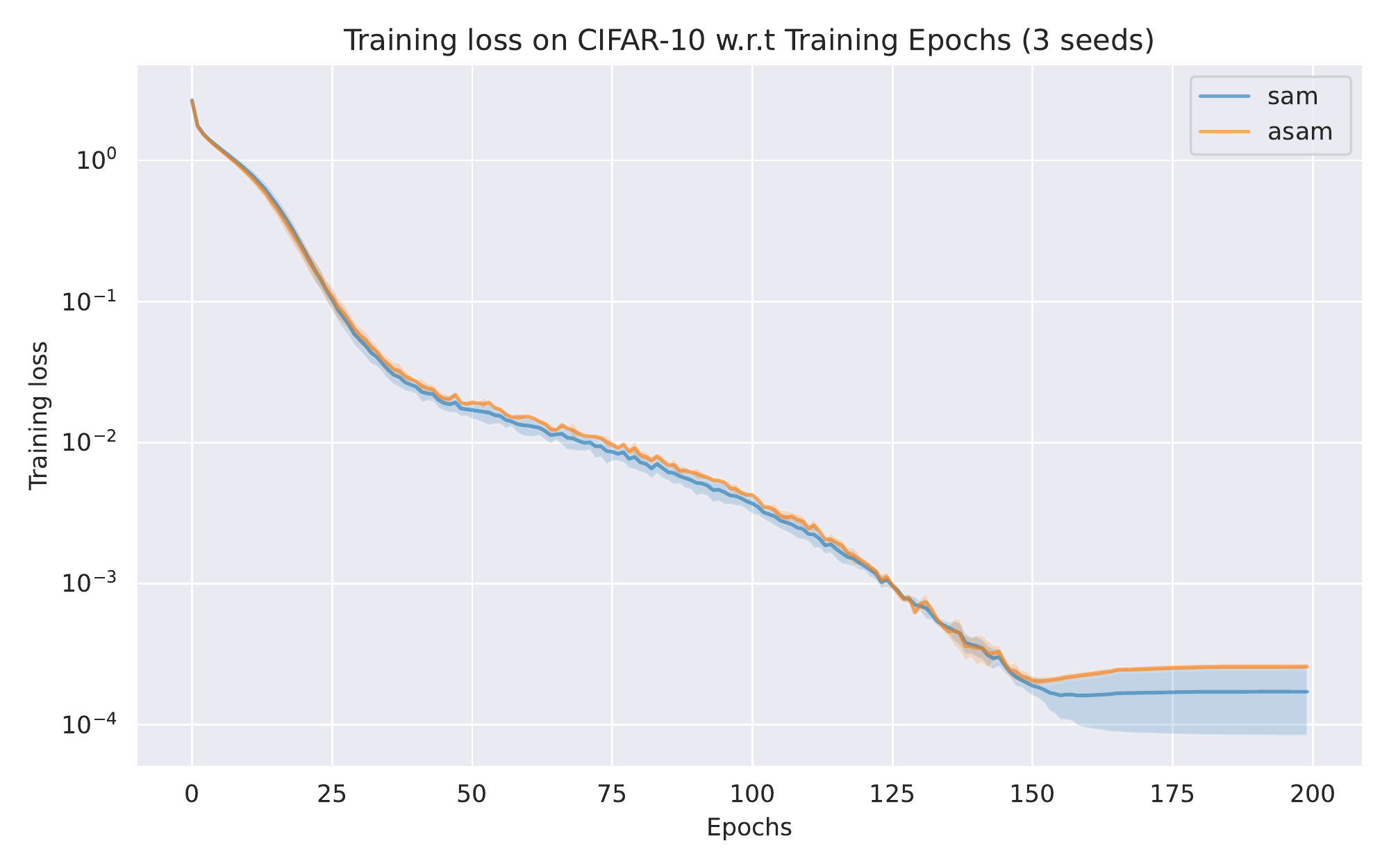}
		\caption{Training loss on CIFAR-10 with SAM/ASAM for different initial weights.}
		\label{fig:asamloss}
	\end{figure}

	\section{Conclusion and Future Work}
	As we have seen in experiment a, using sharpness aware minimization in during training helps to achieve better generalization performance during testing.
	
	In addition, experiment b shows that even though SAM and vanilla SGD both suffer from degradation of generalization performance for larger batch sizes, but SAM mitigates this effect compared to vanilla SGD. In addition, we observed that training sessions with vanilla SGD tend to have higher variance than training sessions with SAM, so we suggest that there may be further correlations between landscape sharpness, gradient variance, and training loss variance, which merit further investigation. A counter-intuitive observation is that low training loss may not always be a good sign in terms of generalization performance.
	
	Lastly, experiment c showed that the removal of the batch normalization resulted in the dirstribution of weights significantly changing. The distribution of weights in the networks without batch normalization showed a much heavier tail, indicating that there is a large difference in the scale of the parameter. We have also seen that the ASAM algorithm generally performs better than the SAM algorithm on this task. Moreover, we saw a larger difference in the variance of the training loss in the SAM algorithm, while ASAM seemed to be much more constant for different weight initializations. Future work could expand on these experiments by testing the difference in performance of SAM and ASAM on more models and different data-sets.

	\newpage
	\bibliographystyle{IEEEtran}
	\bibliography{literature}
	
	\newpage
	\appendix
	\section{Appendix}
	
	\begin{figure}[H]
		\centering
		\includegraphics[width = 0.45\textwidth]{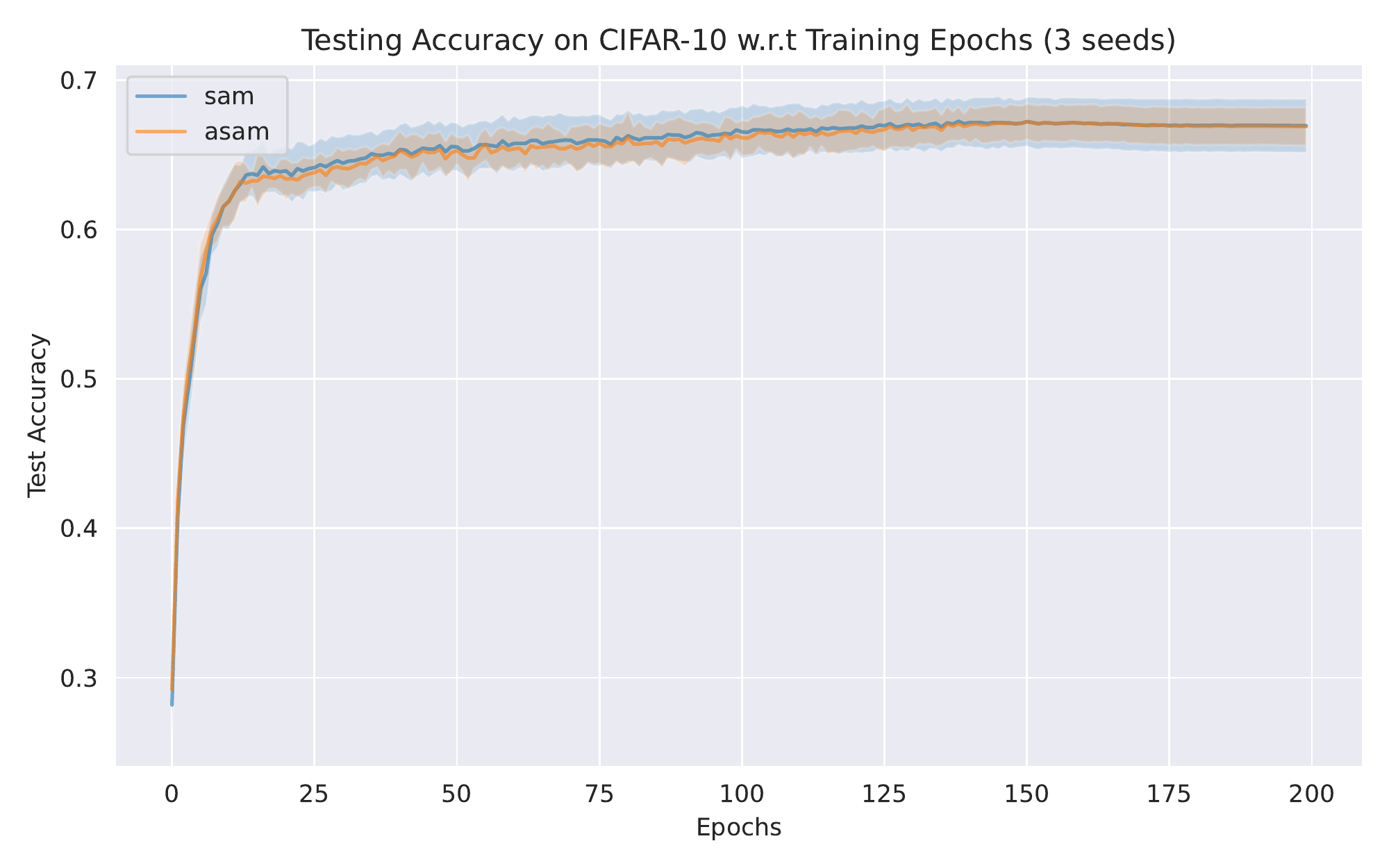}
		\caption{Testing accuracy on CIFAR-10 with SAM/ASAM for different initial weights.}
		\label{appendix:asamacc}
	\end{figure}
	
\end{document}